MIMIC-IV-Ext-PE: Using a large language model to predict pulmonary embolism phenotype in the MIMIC-IV dataset

Barbara D. Lam[1,2], Shengling Ma[3], Iuliia Kovalenko[4], Peiqi Wang[5], Omid Jafari[6], Ang Li[3]\*, Steven Horng[2,7]\*

1. Division of Hematology, Department of Medicine, Beth Israel Deaconess Medical Center, Boston, MA, USA
2. Division of Clinical Informatics, Department of Medicine, Beth Israel Deaconess Medical Center, Boston, MA, USA
3. Section of Hematology-Oncology, Department of Medicine, Baylor College of Medicine, TX, USA
4. Department of Medicine, University of Pittsburgh Medical Center, Harrisburg, PA, USA
5. Computer Science and Artificial Intelligence Laboratory, Massachusetts Institute of Technology, Cambridge, MA, USA
6. Institute for Clinical and Translational Research, Baylor College of Medicine, TX, USA
7. Department of Emergency Medicine, Beth Israel Deaconess Medical Center, Boston, MA, USA

\*co-senior authors

**Abstract**
Pulmonary embolism (PE) is a leading cause of preventable in-hospital mortality. Advances in diagnosis, risk stratification, and prevention can improve outcomes. There are few large publicly available datasets that contain PE labels for research. Using the MIMIC-IV database, we extracted all available radiology reports of computed tomography pulmonary angiography (CTPA) scans and two physicians manually labeled the results as PE positive (acute PE) or PE negative (gold standard). We then applied a previously finetuned Bio_ClinicalBERT transformer language model, VTE-BERT, to extract labels automatically. We verified VTE-BERT's reliability by measuring its performance against manual adjudication. We also compared the performance of VTE-BERT to the use of diagnosis codes. We found that VTE-BERT has a sensitivity of 92.4% and positive predictive value (PPV) of 87.8% on all 19,942 patients with CTPA radiology reports from the emergency room and/or hospital admission. In contrast, diagnosis codes have a sensitivity of 95.4% and PPV of 83.8% on the subset of 11,990 hospitalized patients with discharge diagnosis codes. We successfully added nearly 20,000 labels to CTPAs in a publicly available dataset and demonstrate the external validity of a semi-supervised language model in accelerating hematologic research.

**Background**
Pulmonary embolism (PE) is a leading cause of preventable in-hospital mortality[1]. Early detection and treatment can reduce the risk of death, but diagnosis can be challenging because the typical symptoms of tachycardia, shortness of breath, and chest pain are nonspecific. Much work has been done to identify patients at risk of developing PE and tailor treatment for those who have been diagnosed with PE, but better risk assessment models are needed[2]. Research groups are exploring the use of machine learning techniques to improve PE detection and treatment[3]. External validation of these models in a different healthcare system or dataset is a critical step in advancing the field, but there are few large publicly available datasets with PE labels[4].

Identifying PE in medical charts is not only important for furthering research in thromboembolism but also for public health monitoring. The majority of hospital-acquired PE are thought to be preventable, and reporting the incidence of PE has become an important hospital quality metric[1,5]. Historically, PE diagnoses were identified by manual chart review, which was labor intensive and difficult to scale. Attempts to use International Classification of Diseases (ICD) diagnosis codes revealed poor predictive value, especially in the outpatient and emergency room settings[6,7]. Natural language processing (NLP), a computerized approach to parsing and extracting information from text, potentially offers a more accurate and automated method for identifying diagnoses in the chart[8]. NLP methods can range from manually developed *if these, then that* rule-based algorithms to more modern machine learning techniques. Based on a recent systematic review, very few research groups have completed an external validation of their work and none that we know of have attempted using a transformer language model to identify PE in unstructured medical notes[9].

Transformer language models represent the most recent iteration of NLP and are changing the landscape of numerous fields including healthcare[10]. Maghsoudi et al previously customized the transformer language model known as Bio_ClinicalBERT by finetuning its ability to identify venous thromboembolism (VTE) including deep vein thrombosis (DVT) and PE in a cohort of 800 cancer patients with 3000 notes[11]. This finetuned model is referred to as the VTE-BERT model from here forward. In this study, we reviewed all radiology reports in the publicly available MIMIC-IV dataset and identified those describing a computed tomography scan of the pulmonary arteries (CTPA), which is the gold standard imaging for diagnosing PE. We used VTE-BERT to automatically label the reports as PE positive (acute PE) or negative, and compared its performance to diagnosis codes and manual adjudication by two physicians. Our code and the final labeled CTPA dataset are shared for easy replication and to encourage the external validation of emerging models.

**Methods**
We utilized the MIMIC-IV dataset, which is the most recently released version of the Medical Information Mart for Intensive Care[12]. MIMIC-IV includes all patients at Beth Israel Deaconess Medical Center (BIDMC) who presented to the emergency room or were admitted to the intensive care unit between 2008-2019[13]. It includes detailed clinical data for over 400,000 patients and is one of the most commonly used publicly available datasets for machine learning research around the world. All data in MIMIC-IV has been previously de-identified, and the institutional review boards of the Massachusetts Institute of Technology (0403000206) and BIDMC (2001-P-001699/14) both approved use of the database for research.

We first used Regular Expression (RegEx) to review all notes of subtype "RR" (radiology report) in MIMIC-IV and identified those that are CTPAs. RegEx is a rule-based NLP approach that can identify text based on character patterns and has previously shown to be effective for parsing radiology reports[14]. Reports were segmented into "History", "Indication", "Procedure", "Examination", "Study", and "Technique" sections. Any reports that contained phrases related to

PE in the first two sections and phrases related to CTPAs in the last four sections were included. Reports that contained a separate section describing CTPAs were also included (Supplementary Table 1).

Two physicians (BDL and IK) confirmed that each radiology report described a CTPA. Any type of imaging study that also included a CT angiogram of the chest was included. For example, imaging studies such as CT angiograms of the entire torso or CT angiograms of the cardio-vasculature were included if a CT angiogram of the chest was described in the technique section.

CTPA radiology reports were further preprocessed using an algorithm that identified PE keywords, isolated relevant sentence(s), and merged them into a final note file (Supplementary File 2). These sentences were used as input to VTE-BERT, which was asked to predict whether the compiled notes described an acute PE (PE positive) or not (PE negative). If no sentence was isolated by the preprocessing algorithm for evaluation by VTE-BERT, the prediction was labeled as negative (predicting no PE).

The isolated sentence and VTE-BERT prediction were reviewed by one physician (BDL). The second physician reviewed the isolated sentence only and was blinded to the VTE-BERT output (IK). The physicians reviewed the entire radiology report if there was no isolated sentence. Both physicians used the following gold standard criteria for categorizing PE:

- Positive PE
    - Acute = Acute PE or a mix of acute and chronic PE
    - Subsegmental = Acute PE in subsegmental arteries only
- Negative PE
    - Chronic = Chronic PE, PE of unclear chronicity, or PE similar to last scan
    - Equivocal = Equivocal findings such as motion artifact versus PE
    - Negative = No PE, imaging suboptimal for identifying PE, or no description of the pulmonary arteries

All conflicts were discussed until agreement was reached for every report.

We also assessed the accuracy of using ICD codes to identify PE cases in a subgroup analysis. In MIMIC-IV, the ICD codes associated with radiology reports are only available for hospitalizations with billable discharges. Therefore, we only assessed the performance of ICD codes on radiology reports that had an associated hospital identification number. ICD-10 codes starting with I26 and ICD-9 codes starting with 415 were included as PE-related. I26.01, I26.90, and 415.12 specifically were excluded because they refer to septic PE. One patient could have multiple CTPAs during their hospital stay; if this occurred then only one CTPA report was included for comparison. If one of the reports showed an acute PE, that report was preferentially included. All analyses were conducted using Python 3.11.

**Results**
Of 2,321,355 radiology reports in MIMIC-IV, we identified 21,948 reports as likely CTPAs (Figure 1). After manual review, we confirmed 19,942 distinct CTPA reports from 15,875 patients. The median age was 58 years and approximately half of the patients identified as female (51.3%) and white (59.8%) (Table 1).

Based on manual abstraction (gold standard), among the 19,942 CTPAs identified, 1,591 described acute PE (of which 233 involved subsegmental arteries only) and 18,351 described negative PE (of which 345 described chronic PE and 104 described equivocal findings). Our CTPA positivity rate was 7.98%, which is in the range of other previously reported rates although our findings are difficult to compare because we also included imaging studies that were done for different indications[15]. These final PE labels will be available for download through the MIMIC-IV dataset website.

Our preprocessing algorithm identified and isolated PE-containing relevant sentences from 18,748 reports. Among the remaining 1,194 reports where no relevant keywords were identified, only one described an acute PE. The VTE-BERT model demonstrated a sensitivity of 0.92 (95% CI: 0.91-0.94) and a positive predictive value (PPV) of 0.88 (95% CI: 0.86-0.89) (Table 2). The most common error was prediction of a report to be positive when it described chronic PE findings only.

Among the 19,942 CTPAs identified, 12,355 were associated with 11,990 unique hospital stays (365 represented multiple images from the same stay). When comparing the inpatient discharge ICD codes to the physician-adjudicated gold standard, we found that 308 reports were incorrectly labeled by ICD code: 61 reports described an acute PE that had no relevant ICD code associated; and of those with an ICD code indicating acute PE, 108 reports described chronic PE only, 115 were negative for PE, and 24 were deemed equivocal findings. Of the 1,276 reports that were correctly identified as PE positive by ICD code, 169 described PE involving the subsegmental arteries only. Four of these 169 reports had an ICD code that specified PE in the subsegmental arteries only. Overall, ICD codes demonstrated a sensitivity of 0.95 (95% CI: 0.94-0.96) and a PPV of 0.84 (95% CI: 0.82-0.86) for identifying PE in inpatient visits (Table 2). Due to the limitations of ICD codes for emergency room only visits in MIMIC-IV, 7,587 reports were not analyzed.

**Discussion**
In this study, we successfully externally validated and applied a finetuned transformer language model, VTE-BERT, to label nearly 20,000 CTPA reports in MIMIC-IV as PE positive or PE negative. VTE-BERT demonstrated high sensitivity for predicting which CTPA reports described an acute PE. Notably, no additional transfer learning was done to improve VTE-BERT before testing it on MIMIC-IV radiology reports and thus this is a true external validation of a finetuned transformer language model. Challenges included CTPA reports that described chronic PE only, and work is being done to further tune the model to these examples before it is released to the public.

There are several advantages to using transformer language models over ICD codes for labeling data. These models can be applied to any semi-structured or unstructured free-text in the inpatient, outpatient, and emergency room settings, whereas ICD codes may not be uniformly available. Model performance can be continually improved and finetuned based on human feedback, while ICD codes are mostly static for billing reasons. Finally, the accuracy of ICD codes largely depends on its input source and data availability. Facility billing discharge diagnosis codes that are entered by medical billing professionals may be more accurate, while encounter and professional billing diagnosis codes entered by medical providers may contain more errors. A common practice for researchers relying on ICD codes for phenotyping is to use singular inpatient ICD codes or two or more outpatient ICD codes more than 30 days apart, a practice that lends significant complexity to population health sciences research[16].

This study demonstrates the power of transformer language models in labeling large amounts of data with minimal human intervention. The underlying architecture of VTE_BERT is BERT, a transformer language model that can be readily downloaded and customized to specific tasks. Newer iterations of transformer language models, often referred to as large language models (LLMs), offer different strengths and weaknesses. LLMs have increased capacity, which has been associated with improved performance on tasks[17]. However, the largest models tend to be proprietary in nature and researchers cannot upload protected health information to models on the web without an institutional agreement in place. There are smaller models including those trained on medical text specifically that are open access and available for download and use, but finetuning these models will require significant computational resources that not all researchers have access to[18]. LLMs are known for their generalizability and may be able to perform data labeling tasks through curation of the input alone (a process known as "prompt engineering") without any explicit training or change to the underlying architecture[19]. Further research is needed to explore how different types of language models can best be leveraged to label large datasets and accelerate research.

ICD codes demonstrated a sensitivity of 95.4% and 83.8% PPV for identifying PE in the chart, but in this dataset could only be applied to radiology reports associated with an inpatient stay; 7,587 CTPAs could not be evaluated using diagnosis codes and likely had lower concordance. As reported previously in the literature, ICD codes have poor predictive value for identifying PE, especially in the emergency room and outpatient settings[6,7]. Furthermore, only four hospital stays had an ICD code that specified PE in the subsegmental arteries only. This level of classification is important for research given controversy around the clinical relevance of subsegmental PE[20,21]. One limitation of our evaluation of ICD codes, however, is the lack of insight into the patient's clinical presentation at the time of the radiology study. For example, a patient with equivocal findings on the radiology report but significant symptoms may have ultimately been treated for PE and therefore labeled with an ICD diagnosis code of PE. Future work can be done to investigate which patients were treated with anticoagulation as further validation of the PE diagnosis on the patient or hospital stay level.

Adding nearly 20,000 PE labels to CTPA reports in MIMIC-IV opens the door for this type of future work and other large-scale studies. The patient and hospital stay identifiers for each report can be linked to tabular data including vital signs, laboratory studies, and administered medications, free-text data such as nursing notes, and imaging studies such as electrocardiograms. Our work adds a large, publicly available multimodal dataset for PE to the literature, which is critical for the expansion of machine learning research in hemostasis and thrombosis[4]. Information on optimal PE risk stratification, diagnosis, and treatment lies in various types of data and these large datasets enable research into multimodal approaches to PE management[22]. They also enable researchers to easily test the performance of their models on external data, an important validation step prior to clinical deployment. The code we share to identify CTPA scans and isolate the relevant sentence describing PE can also be adapted to identify and label other types of studies in large datasets, for example, adding DVT labels to duplex ultrasounds.

Our PE labels were confirmed by dual physician adjudication. It is possible that the physician reviewer could be biased by the VTE-BERT prediction, particularly in cases where the findings were equivocal. We attempted to minimize this bias by blinding the second physician reviewer to the VTE-BERT prediction. However, there can still be human errors in manual adjudication and subjectivity in interpreting radiology reports. We utilized the label of equivocal findings to flag the CTPA reports with less definitive language. We invite others to replicate our work and iterate on the final dataset, which can undergo continued improvements in the public space.

**Conclusion**
Transformer language models are a promising tool for curating datasets, which can support the development and external validation of models in the field of hemostasis and thrombosis.

**Data sharing**

The PE labels derived from this study will be shared through the MIMIC-IV website.
Please refer to the corresponding GitHub repository for relevant code:
1. [RegEx algorithm for identifying CTPA studies from all radiology reports](#)
2. [RegEx preprocessing algorithm for identifying relevant PE sentence](#)

**Figures and tables**
**Figure 1. Study flowchart**

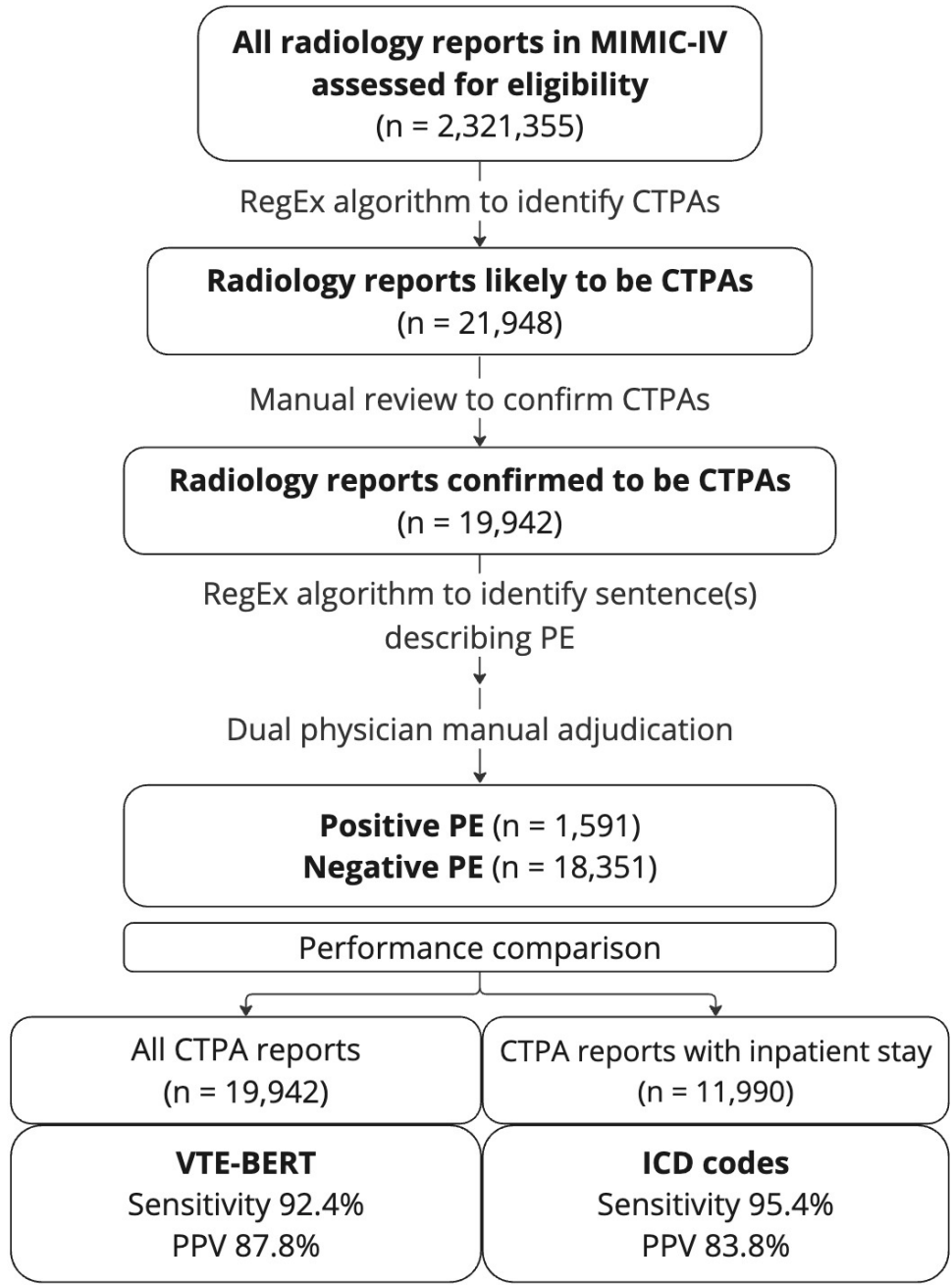

**Table 1. CTPA radiology report demographics**

|  | CTPA radiology reports n (%) |
|---|---|
| **Radiology reports** | 19,942 |
|   Associated with inpatient hospitalization | 12,355 (62.0%) |
|   Associated with emergency room visit only | 7,587 (38.0%) |
|   Acute PE | 1,591 (8.0%) |
|     Subsegmental arteries only | 233 (1.2%) |
|   Chronic PE | 345 (1.7%) |
|   Equivocal findings | 104 (0.5%) |
| **Unique patients** | 15,875 |
|   Median age | 58 years |
|   Female | 8,148 (51.3%) |
|   Race/Ethnicity |  |
|     White | 9,500 (59.8%) |
|     Black | 2,553 (16.1%) |
|     Asian | 477 (3.0%) |
|     American Indian/Alaska Native | 22 (.1%) |
|     Hispanic/Latino | 830 (5.2%) |
|     Other/Unknown* | 2,493 (15.7%) |

*This category includes other, unknown, multiple race/ethnicity, Portuguese, South American, patient declined to answer, unable to obtain

**Table 2. Two-by-two tables demonstrating performance of VTE-BERT on all CTPA reports, ICD codes on CTPA reports with an inpatient stay, and VTE-BERT on discharge summaries**

|  | True acute PE based on physician review of CTPA report | True negative based on physician review of CTPA report |  |
|---|---|---|---|
| **VTE-BERT language model (n=19,942)** | | | |
| VTE-BERT for CTPA report predicts PE positive | 1,470 | 204 | PPV = 87.8% |
| VTE-BERT for CTPA report predicts PE negative | 121 | 18,147 | NPV = 99.3% |
|  | Sensitivity = 92% | Specificity = 98.9% |  |
| **ICD-CM discharge diagnosis codes in subset of hospitalized patients (n=11,990)*** | | | |
| ICD codes for acute PE present at discharge | 1,276 | 247 | PPV = 83.8% |
| ICD codes for acute PE absent at discharge | 61 | 10,406 | NPV = 99.4% |
|  | Sensitivity = 95.4% | Specificity = 97.7% |  |

*Performance of ICD codes for CTPAs associated with an emergency room visit only (n=7,952) could not be analyzed due to limitations of dataset

# Supplementary materials

## Supplementary Table 1: Terms used to identify CTPA radiology reports in MIMIC-IV

| Section of radiology report | Procedure, Examination, Study, Technique | History, Indication | CTPA |
|---|---|---|---|
| Terms that were used to determine which radiology reports to include | CTA chest<br>CTA pulmonary angiogram<br>CTA of the chest<br>Chest CTA<br>CTPA<br>Torso CTA<br>CTA torso | Pulmonary embolus<br>Pulmonary emboli<br>Pulmonary embolic<br>Pulmonary embolism<br>Pulmonary thromboembolism<br>Pulmonary artery embolus<br>Pulmonary artery emboli<br>Pulmonary artery embolic<br>Pulmonary artery embolism<br>Pulmonary artery thromboembolism<br>Pulmonary arterial embolus<br>Pulmonary arterial emboli<br>Pulmonary arterial embolic<br>Pulmonary arterial embolism<br>Pulmonary arterial thromboembolism | CTA chest<br>CTA of the chest<br>CTA thorax |
| Number of radiology reports that met inclusion criteria | N=17,386 | N=2,707 | N=2,077 |

CTPA = computed tomography pulmonary angiogram; CTA = computed tomography anigiogram

## Supplementary File 2: Terms used in RegEx preprocessing algorithm

| PE-like keywords that were identified for exclusion |
|---|
| PE CT, PECT, PE-CT, PE/CT, CT PE, CT/PE, CT-PE, DVT US, DVT U/S, DVT ultrasound, PE protocol, PE study, PE technique, PE scan, DVT protocol, DVT study, DVT technique, DVT scan, VTE prophylaxis, VTE prophy, VTE ppx, DVT prophylaxis, DVT prophy, DVT ppx |
| PE keywords that were identified for inclusion |
| PE, VTE, pulmonary embolus, pulmonary emboli, pulmonary embolic, pulmonary embolism, pulmonary thromboembolism, pulmonary artery embolus, pulmonary artery emboli, pulmonary artery embolic, pulmonary artery embolism, pulmonary artery thromboembolism, pulmonary arterial embolus, pulmonary arterial emboli, pulmonary arterial embolic, pulmonary arterial embolism, pulmonary arterial thromboembolism, thromboemboli, thromboembolism, filling defect, filling defects, embolus, emboli, embolic, embolism, embolisms |